\documentclass[10pt,twocolumn,letterpaper]{article}

\usepackage{cvpr}              

\usepackage{times}
\usepackage{epsfig}
\usepackage{graphicx}
\usepackage{amsmath}
\usepackage{amssymb}
\usepackage{booktabs} 
\usepackage{algorithm}
\usepackage{algorithmic}
\usepackage{amsthm}
\usepackage{diagbox}
\usepackage{caption}
\usepackage{float}
\usepackage{epstopdf}
\usepackage{enumitem}
\usepackage{multirow}
\usepackage{bbm}
\usepackage[dvipsnames]{xcolor}

\usepackage{amsmath,amsfonts,bm}

\def\eqref#1{equation~\ref{#1}}

\def\1{\bm{1}}

\def\va{{\bm{a}}}

\def\vx{{\bm{x}}}

\def\mA{{\bm{A}}}

\def\mW{{\bm{W}}}
\def\mX{{\bm{X}}}

\DeclareMathAlphabet{\mathsfit}{\encodingdefault}{\sfdefault}{m}{sl}
\SetMathAlphabet{\mathsfit}{bold}{\encodingdefault}{\sfdefault}{bx}{n}

\newcommand{\R}{\mathbb{R}}

\newcommand{\softmax}{\mathrm{softmax}}

\definecolor{citecolor}{HTML}{0071bc}
\usepackage[pagebackref=false,breaklinks=true,letterpaper=true,colorlinks,citecolor=citecolor,bookmarks=false]{hyperref}

\usepackage[capitalize]{cleveref}
\crefname{section}{Sec.}{Secs.}
\Crefname{section}{Section}{Sections}
\Crefname{table}{Table}{Tables}
\crefname{table}{Tab.}{Tabs.}

\usepackage{color,colortbl}
\definecolor{ballblue}{rgb}{0.13, 0.67, 0.80}
\definecolor{amaranth}{rgb}{0.90, 0.17, 0.31}
\definecolor{olive}{rgb}{0.5, 0.5, 0.0}
\definecolor{Gray}{gray}{0.85}
\newcolumntype{a}{>{\columncolor{Gray}}c}
\definecolor{blush}{rgb}{0.87, 0.36, 0.51}
\newcommand{\cm}{\texttt{COMEN}}

\def\cvprPaperID{157} 
\def\confName{CVPR}
\def\confYear{2022}

\begin{document}

\title{Compound Domain Generalization via Meta-Knowledge Encoding}

\author{Chaoqi Chen$^1$, Jiongcheng Li$^{2,4}$, Xiaoguang Han$^{3}$, Xiaoqing Liu$^2$, Yizhou Yu$^{1}\thanks{Corresponding author}$\\
	{$^1$~The University of Hong Kong}\quad{$^2$~Deepwise AI Lab}\\
	{$^3$~The Chinese University of Hong Kong (Shenzhen)}\quad$^4$~{Xiamen University}\\
	{\tt\small cqchen1994@gmail.com, jiongchengli@stu.xmu.edu.cn}\\
	{\tt\small hanxiaoguang@cuhk.edu.cn, Liuxiaoqing@deepwise.com, yizhouy@acm.org}
}
\maketitle

\begin{abstract}
Domain generalization (DG) aims to improve the generalization performance for an unseen target domain by using the knowledge of multiple seen source domains. Mainstream DG methods typically assume that the domain label of each source sample is known a priori, which is challenged to be satisfied in many real-world applications. In this paper, we study a practical problem of compound DG, which relaxes the discrete domain assumption to the mixed source domains setting. On the other hand, current DG algorithms prioritize the focus on semantic invariance across domains (one-vs-one), while paying less attention to the holistic semantic structure (many-vs-many). Such holistic semantic structure, referred to as meta-knowledge here, is crucial for learning generalizable representations. To this end, we present \textbf{CO}mpound domain generalization via \textbf{M}eta-knowledge \textbf{EN}coding~\emph{(\cm)}, a general approach to automatically discover and model latent domains in two steps. Firstly, we introduce Style-induced Domain-specific Normalization (SDNorm) to re-normalize the multi-modal underlying distributions, thereby dividing the mixture of source domains into latent clusters. Secondly, we harness the prototype representations, the centroids of classes, to perform relational modeling in the embedding space with two parallel and complementary modules, which explicitly encode the semantic structure for the out-of-distribution generalization. Experiments on four standard DG benchmarks reveal that \emph{\cm} exceeds the state-of-the-art performance without the need of domain supervision.
\end{abstract}

\section{Introduction}
\label{sec:intro}
\begin{figure}[!t]
	\centering
	\includegraphics[width=0.48\textwidth]{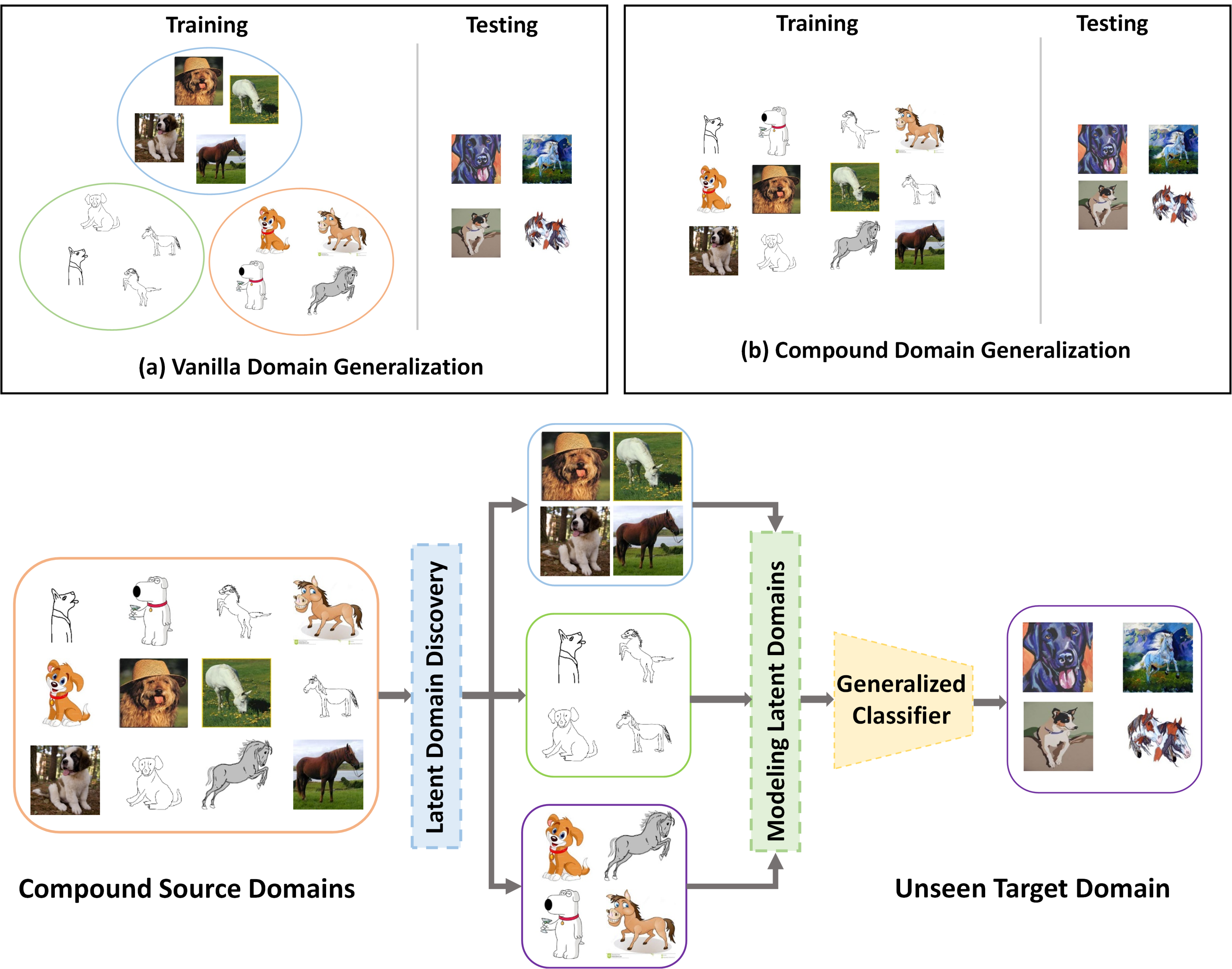}
	\vspace{-0.6cm}
	\caption{\textbf{Top:} (a) Vanilla domain generalization. (b) Compound domain generalization. \textbf{Bottom:} The workflow of our approach.}\label{fig1}
	\vspace{-0.4cm}
\end{figure}

The success of many computer vision algorithms hinges on a strong presumption that the training and test data are independent and identically distributed (i.i.d.).
In practice, however, this hypothesis is prone to be violated due to the change of environments, imaging devices, selection bias, to name a few. 
How to generalize a well-trained model to out-of-distribution domains has motivated a body of research on Domain Adaptation (DA)~\cite{tzeng2014deep,long2015learning,ganin2016domain,sun2016deep,saito2018maximum,zhang2019domain,zhang2020unsupervised} and Domain Generalization (DG)~\cite{motiian2017unified,li2018domain,carlucci2019domain,li2019episodic,dou2019domain,chen2022self}.
In contrast to DA problem where the unlabeled or partially labeled target data is available, 
DG considers a harder problem setting where a model trained on a set of source domain data should directly generalize to an unseen target domain with different data statistics, without the need of accessing to target domain data for retraining or fine-tuning (cf. Fig.~\ref{fig1}~(a)). 

\label{introduction}

Mainstream DG literature typically assumes that the domain label of each source sample is known a \emph{priori}, which is too restrictive to be satisfied in many practical applications. 
For instance, with the explosive increase of the multi-source data from Internet, a mixture of multiple source data can be easily collected. 
However, manually dividing the crawled images into multiple domains requires tedious labor-intensive work given a large amount of training data. Meanwhile, the latent domains may blend and interact in complex ways. 
Thus, it is non-trivial and significant to automatically discover and model distinct underlying domains.


In this paper, we investigate a practical problem of \emph{compound domain generalization}, where the domain label is unavailable for training (cf. Fig.~\ref{fig1}~(b)).
An intuitive approach is to use the embedded features along with clustering strategies to conduct domain separation in advance. 
However, the source samples are naturally divided into different clusters according to their semantic categories rather than the domain-wise distinctions. 
When the source data contains heterogeneous but unknown latent domains, most conventional DG methods will not be applicable. 
Although some recent works~\cite{carlucci2019domain,huang2020self,nam2021reducing,li2021simple} do not require domain label information during training, the informative latent domain structure contained in the source data is neglected.

On the other hand, existing DG methods, which mainly span domain invariant feature learning~\cite{muandet2013domain,ghifary2015domain,motiian2017unified,li2018domain,li2018deep,zhang2022exact}, gradient based meta-learning~\cite{li2018learning,li2019episodic,dou2019domain}, and augmentation based generalization~\cite{shankar2018generalizing,volpi2018generalizing}, are devoted to learn semantic representations in virtue of one-vs-one consistency constraints. 
Unfortunately, the learned representations may be insufficient to precisely encode the semantic information. 
By prioritizing the focus on semantic invariance across domains, the holistic semantic structure, which contains rich many-vs-many information regarding the inter-class relations and interactions, is yet to be thoroughly investigated.
In particular, we refer such holistic semantic structure as \emph{meta-knowledge}, namely, knowledge that is domain-agnostic and generalizable for unseen target domains.

Grounded on these findings, we present a novel DG framework called \textbf{CO}mpound domain generalization via \textbf{M}eta-knowledge \textbf{EN}coding~(\cm). 
The basic idea of the \cm\ is to identify latent domain structure and model the semantic correlations among different categories across domains.
To achieve this goal, we establish the \cm\ in two stages. 
During the first stage, we introduce Style-induced Domain-specific Normalization (SDNorm) 
to reveal the complex combinations of latent domains and learn a set of domain assignment probabilities for each source sample. 
SDNorm is end-to-end trainable and can be easily plugged into modern deep neural networks.
In the second stage, based on the discovered latent domains, we resort to prototypes (the feature centroids of categories) to model the relations among different semantic categories via two parallel modules, \emph{i.e.,} Prototypical Graph Reasoning (ProtoGR) and Prototypical Category-aware Contrastive Learning~(ProtoCCL). 
ProtoGR helps each categorical prototype attend and reason over its neighborhoods' prototypical features instead of using pairwise alignment, thereby capturing the topological structure of semantic space. 
ProtoCCL compensates for the shortage of samples in the prototypical feature space by contrastively learning the relations of different categories while preserving their discriminability. 
These two modules work in a synergistic manner towards encoding semantic structures into the embedding space.

Our contributions can be summarized as follows:
\begin{itemize}
	\item{We introduce a more practical compound DG setting that imposes no prior knowledge on the domain label of each source sample. 
		Then, a unified learning framework, called \cm, is introduced to jointly discover and model the latent domains.}
	
	\item{In terms of latent domain discovery, we uncover and re-normalize the multimodal underlying distributions with the proposed SDNorm.}
	
	\item{In terms of encoding the semantic structure, we propose two complementary modules, ProtoGR and ProtoCCL, to explicitly explore the relations and interactions of prototype in the common feature space.}
	
	\item{Experiments on four standard DG benchmarks (PACS, Digits-DG, VLCS, and Office-Home) demonstrate that \cm\ outperforms the state-of-the-art methods by large margins without the need of domain supervision.}
\end{itemize}

\begin{figure*}[!t]
	\centering
	\includegraphics[width=1\textwidth]{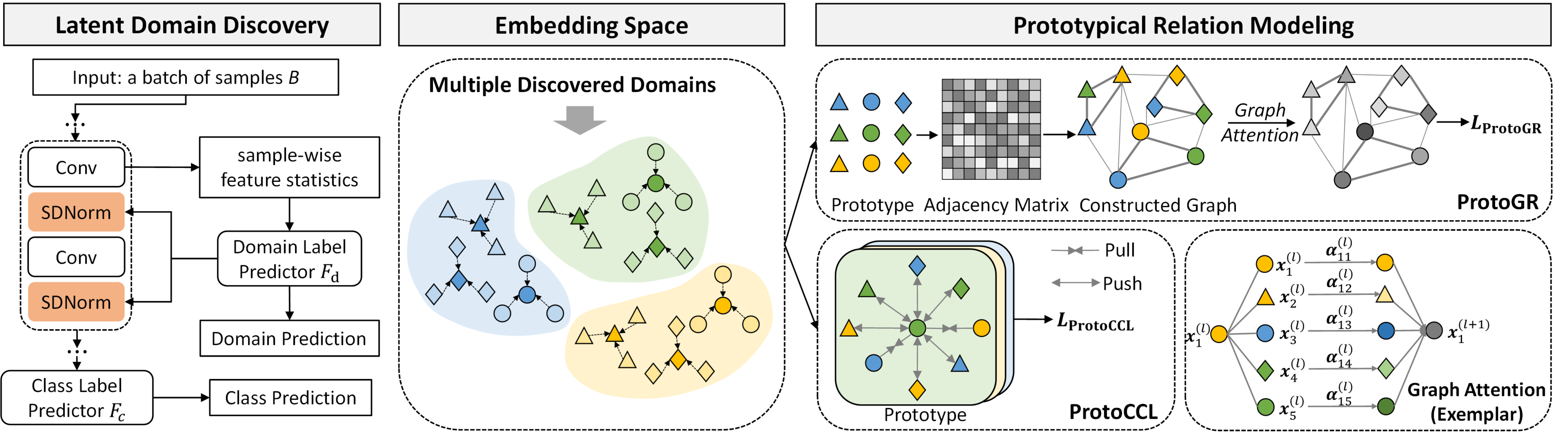}
	\caption{The overall structure of \cm. Given a mixture of source domains, it discovers the latent domain structure via SDNorm in the first stage. During the second stage, it first converts all samples into the embedding space with pseudo domain labels (based on the domain prediction). Then, a set of prototypes are computed and fed into two parallel modules, ProtoGR and ProtoCCL.}
	\label{fig2}
\end{figure*}

\section{Related Work}
\label{sec:related_work}
\subsection{Domain Generalization}
Domain generalization~\cite{wang2021generalizing}, which aims at generalizing model learned from multiple seen source domains to an unseen target domain, have received a great surge of interest in the deep learning era. In practice, numerous approaches have been developed to learn a good predictive model for the unseen test domain with generalizable representations. 

\textbf{Domain Alignment.} An intuitive solution for domain generalization is to extract features that are domain-invariant but preserve discriminability for the task prediction. For example, Ghifary~\emph{et al.}~\cite{ghifary2015domain} propose to extract invariant features with a multi-task auto-encoder.  
Motivated by the success of mainstream domain adaptation methods, domain adversarial training~\cite{ganin2016domain} are introduced to adversarially learn generalizable representations across multiple source domains via a two-player min-max game~\cite{li2018deep,li2018domain,dg_mmld}.

\textbf{Augmentation.} Volpi~\emph{et al.}~\cite{volpi2018generalizing} augment the dataset with a new adversarial data augmentation procedure by synthesizing the ``hard'' samples for the training model, and then provide an an iterative updating procedure. 
Xu~\emph{et al.}~\cite{xu2021robust} resort to randomized convolutions by leveraging the outputs of random convolutions as new images or mixing them with the original inputs during training.
By contrast, feature augmentation, such as mixstyle~\cite{zhou2021domain}, curriculum mixup~\cite{mancini2020towards}, and stochastic feature augmentation~\cite{li2021simple}, is newly investigated to augment the latent feature space without explicitly changing the original inputs.
In addition, Carlucci~\emph{et al.}~\cite{carlucci2019domain} use a self-supervision loss, called solving jigsaw puzzles, to enhance the generalization ability of deep model by distinguishing different jigsaw puzzles. 

\textbf{Meta-Learning.} A category of methods resort to the meta-learning~\cite{finn2017model} by constructing meta-train and meta-test domains during training time for mimicking the train and test distributional disparity. Li~\emph{et al.}~\cite{li2018learning} and Balaji~\emph{et al.}~\cite{balaji2018metareg} propose to update the network parameters by minimizing the meta-optimization objective. 
Li~\emph{et al.}~\cite{li2019episodic} first develop a strong baseline by aggregating data from all source domains and training a deep model on all the data, and then a episodic training strategy is proposed to mimic the domain shift. 
Dou~\emph{et al.}~\cite{dou2019domain} propose to explicitly enforce the cross-domain semantic coherence via metric learning losses along with the episode training strategy to enhance the semantic robustness. Despite their general efficacy, meta-learning methods are notoriously slow to train.

\textbf{Feature Decomposition.} Khosla~\emph{et al.}~\cite{khosla2012undoing} propose to a decomposition-based approaches that decompose network parameters to domain-invariant and domain-specific low-rank components. This idea was further extended by~\cite{li2017deeper} to the deep learning model. 
A low common-specific low-rank decomposition algorithm~\cite{piratla2020efficient} is proposed to achieve efficient domain generalization by only modifying the final linear classification layer of deep networks. 

\subsection{Multiple Latent Domains}
Learning from multiple latent domains is a long-standing and challenging problem in machine learning and computer vision community. 
Some previous works have explored the latent domain discovery problem, but most in the regime of domain adaptation.
In particular, 
Hoffman~\emph{et al.}~\cite{hoffman2012discovering} develop a hierarchical clustering technique to achieve domain separation for the following adaptation procedure.  
Gong~\emph{et al.}~\cite{gong2013reshaping} aim to reshape the visual datasets by imposing two properties (distinctiveness and learnability) on domains.
Xiong~\emph{et al.}~\cite{xiong2014latent} draw motivation from multiple manifold learning to infer the domain assignment by a new local subspace representation.
Mancini~\emph{et al.}~\cite{mancini2018boosting} propose to automatically discover latent domains by using a set of mDA-layers to estimate the domain assignment of each source sample.
In contrast, how to discover and model latent domains are less investigated in DG especially under the presence of unseen target domains.
More importantly, exploring the topological relations among multiple latent domains remains out of reach for current DG approaches. 

\section{Methodology}
The overall architecture of \cm\ is depicted in Fig.~\ref{fig2}, 
which consists of three components, SDNorm, ProtoGR, and ProtoCCL. 
SDNorm statistically estimates the domain membership of each sample in an unsupervised fashion to decouple the multi-modal data distributions. 
ProtoGR and ProtoCCL are built on the basis of prototype representations to model the both intra- and inter-domain semantic relations among different categories, thereby encoding the semantic structure into the learned feature spaces.
\subsection{Task Formulation}
In vanilla domain generalization problem, we are given $M$ source domains $\mathcal{D}_s=\{\mathcal{D}_1,\mathcal{D}_2,...,\mathcal{D}_M\}$ that are sampled from different probability distributions on the joint space $\mathcal{X}\times\mathcal{Y}$. 
It requires the presence of domain labels, thus the $m$-th source domain can be represented as $\mathcal{D}_m=\{(x_i^{m},y_i^{m},d_i^{m})\}_{i=1}^{N_m}$, where $N_m$ denotes the number of samples in this domain. In our work, we focus on the compound domain generalization problem which assumes that the domain labels of source data are not available, \emph{i.e.,} the source data has mixed domains $\mathcal{D}_s=\{(x_i,y_i)\}_{i=1}^{N}$, where $N$ denotes the number of all source samples. 
Both vanilla and compound domain generalization assume that different source domains share the identical label space, and we consider that there are $K$ classes in all.


\subsection{Latent Domain Discovery}
A strong baseline~\cite{li2017deeper} for compound (or vanilla) DG is to train a prediction model by simply combining all source samples. 
However, this approach inevitably deteriorates the performance on the domain of interest~\cite{torralba2011unbiased} as the intrinsic inter-domain relationships are not considered. 
This phenomenon indicates that the underlying multi-modal information is crucial for the success of DG methods.
On the other hand, discovering the latent domains in the context of DG remains an open question since the typical representation learning method are naturally devised to divide the input data into different semantic categories, and most existing DG approaches require the domain information to implement their frameworks. 
In addition, using embedded representations along with standard clustering approaches, such as $k$-means, 
may not lead to satisfactory estimation results with respect to the domain membership of each sample. 
Remedying this issue, we propose to uncover the multi-modal distribution patterns with a simple and effective Style-induced Domain-specific Normalization (SDNorm) module, which can be easily plugged into modern deep neural networks and DG methods.

Suppose that we are given feature maps of an image extracted from a CNN layer $\bm{f} \in \mathbb{R}^{C \times H \times W}$, where $C$, $H$ and $W$ represent the channel, height, and width of the feature maps. 
Technically, we resort to the channel-wise mean and standard deviation $\mu$, $\sigma$ of $\bm{f}$ to obtain its style representation.
The definition is formulated as,
\begin{equation} \label{eq:instance_mean}
	\small
	\mu(\bm{f}) = \frac{1}{HW} \sum_{h=1}^H \sum_{w=1}^W f_{h,w}
\end{equation}
\begin{equation} \label{eq:instance_std}
	\small
	\sigma(\bm{f}) = \sqrt{ \frac{1}{HW} \sum_{h=1}^H \sum_{w=1}^W ( f_{h,w} - \mu(\bm{f}))^2 + \epsilon}
\end{equation}
Then, we concatenate the computed $\mu(\bm{f})$ and $\sigma(\bm{f})$ in each channel to represent the style representation, $\text{sty}(\bm{f})\in\mathbb{R}^{2C}$.
The style representation $\text{sty}(\bm{f})$ is further fed into a domain label predictor $F_d$~\footnote{$F_d$ was pretrained on a portion of samples with pseudo domain labels.} to output a set of domain assignment probabilities. 
Formally, we optimize this domain prediction module by minimizing the entropy of its output,
\begin{equation} \label{eq:domain_prediction}
	\small
	\mathcal{L}_d = -\frac{1}{N}\sum\limits_{i=1}^N\;F_d(\text{sty}(\bm{f})_i)\log{F_d(\text{sty}(\bm{f})_i)}
\end{equation}

We further incorporate the computed domain assignment probabilities into Batch Normalization (BN)~\cite{ioffe2015batch} layer to re-normalize the multi-modal distributions.
BN is devised to standardize the features of each layer to zero-mean and univariance. 
Suppose that we are given an input feature map $\bm{z}$ 
in a certain channel to a BN layer, 
then BN layer transforms the feature by incorporating the domain information in the following manner,
\begin{equation}\label{eq:SDNorm}
	\small
	\hat{\bm{z}}_m=\frac{\bm{z}-\mu_m}{\sqrt{\sigma_m^2+\epsilon}}
\end{equation}
where $\mu_m=\sum\limits_{i=1}^{|B|}\;p_{i,m}\bm{z}^i$ and $\sigma_m^2=\sum\limits_{i=1}^{|B|}\;p_{i,m}(\bm{z}^i-\mu_m)^2$.
$p_{i,m}$ represents the predicted probability from $F_d$ that an input image $x_i$ belongs to the $m$-th domain.
Accordingly, the proposed SDNorm can be formulated as a BN-like module,
\begin{equation}\label{eq:SDNorm}
	\text{SDNorm}_m(\bm{z};\lambda_m,\beta_m)=\lambda_m \cdot \hat{\bm{z}}_m + \beta_m
\end{equation}
where $\lambda_m$ and $\beta_m$ are learnable parameters for domain $m$. 
We perform SDNorm for different source domains by using different bias/variance terms by following the prior practices~\cite{li2017revisiting,carlucci2017autodial,mancini2018boosting,chang2019domain,wang2019transferable,seo2020learning} in domain adaptation. 
Note that we also normalize each channel independently in each mini-batch, but we omit the subscript for the sake of simplicity.

We highlight that the latent domain discovery is an unsupervised process, thus the false domain assignment will inevitably arise during training. Different from representative latent domain discovery methods~\cite{hoffman2012discovering,gong2013reshaping,dg_mmld}, which assign a deterministic domain label (\emph{i.e.} hard domain label) to each source sample, our approach computes a set of probabilities (\emph{i.e.} soft domain label) for each source sample. In this way, the negative influence of false domain assignment will be mitigated during the feature re-normalization process. 

\subsection{Prototypical Relation Modeling}
Mainstream DG approaches are devoted to enforce semantic consistency in the shared embedding space via alignment-based strategies, such as moment matching~\cite{muandet2013domain,sun2016deep} and adversarial learning~\cite{li2018deep,dg_mmld}. However, the semantic relations among different categories, which are crucial for depicting the semantic structures, are less investigated. As a consequence, only seeking one-vs-one category alignment cannot ensure learning generalizable representations for the unseen target domain. On the other hand, \emph{prototype} has demonstrated strong efficacy in few-shot learning~\cite{snell2017prototypical}, domain adaptation~\cite{xie2018learning,chen2019progressive,zhang2020label,chen2021dual}, and unsupervised learning~\cite{li2021prototypical} by encoding semantic structure under the circumstances of insufficient or noisy annotated data. 
In DG, we encounter with a new challenge that how to generalize a learned semantic structure to the unseen domain in an unsupervised fashion. 
To solve the above issues, we propose to model the \emph{prototypical relations} among different semantic categories via two parallel and complementary modules called Prototypical Graph Reasoning (ProtoGR) and Prototypical Category-aware Contrastive Learning~(ProtoCCL). 
ProtoGR seeks to help each category prototype attend and reason over its neighborhoods' prototypical features instead of using pairwise alignment, thereby capturing the topological information of semantic structure. 
ProtoCCL compensates for the limited sample size in the prototypical feature space by contrastively learning the domain relationships and preserving the category discriminability. 

First of all, we compute the global prototype $c_m^{k}$ for each category in different discovered latent domains. 
The global prototype is a mean vector of the embedded samples in each category via a mapping function $f$, 
\begin{equation}\label{eq:global-prototype}
	c_m^{k}=\frac{1}{|\mathcal{D}_m^k|}\sum\limits_{(x_i^{\mathcal{D}_m},y_i^{\mathcal{D}_m})\in{\mathcal{D}_m^k}}f(x_i^{\mathcal{D}_m}),
\end{equation}
where $\mathcal{D}_m^k$ represents the set of samples labeled with class $k$ in the $m$-th source domain. 
In order to harmonize the training process, we follow the prior practice~\cite{xie2018learning,chen2021i3net} to update the global class prototype by using a mini-batch sampled from all source domains through an exponential moving average strategy,
\begin{equation}\label{eq:update-prototype}
	c_{m(I)}^{k}\leftarrow{\rho\,{c_{m(I-1)}^{k}}+(1-\rho)\,\hat{c}_{m(I)}^{k}}
\end{equation}
where $\rho$ denotes the exponential decay rate and is set to 0.7 in all experiments, $\hat{c}_m^{k}$ is the local class prototype, and $I$ represents the iteration times.
\subsubsection{Prototypical Graph Reasoning (ProtoGR)}
To implement ProtoGR, we utilize a graph $\mathcal{G}=\{\mathcal{V},\mathcal{E}\}$ with nodes $v_i \in \mathcal{V}$ and edges $(v_i, v_j) \in \mathcal{E}$. 
In this graph, $\mathcal{V}$ corresponds to $M \times K$ global class prototypes, and the adjacency matrix is defined as $\mA\in\R^{MK \times MK}$ which corresponds to each edge $(v_i, v_j)$ with its element $A_{ij}$.
The node features are denoted by $\mX=\{\vx_1,\cdots,\vx_{MK}\}\in\R^{MK \times d}$, where each element corresponds to a node in $\mathcal{V}$ and $d$ is the dimension of the prototype feature. 
Specifically, we define the correlations between nodes,
\begin{equation}\label{eq:affinity}
	A_{ij}=\mathbbm{1}(\frac{\vx_i^T\vx_j}{{\lVert \vx_i \rVert}_2\cdot{\lVert \vx_j \rVert}_2}>\delta)\cdot{\frac{\vx_i^T\vx_j}{{\lVert \vx_i \rVert}_2\cdot{\lVert \vx_j \rVert}_2}}
\end{equation}
where $\vx_i$ and $\vx_j$ denote the $i$-th and $j$-th prototype features, and $\delta$ is the threshold parameter that control the sparsity of  $\mathcal{G}$. We fix $\delta$ to 0.5 in all experiments. $\mathbbm{1}(\cdot)$ represents the indicator function. 

To model the long-range dependencies and enhance the expressive power of graph nodes, we introduce graph attention mechanism~\cite{velivckovic2018graph} to our model, 
\begin{equation}\label{eq:GAT}
	\vx_i^{(l+1)}=\sigma(\sum_{j \in \mathcal{N}_i}\alpha_{ij}^{(l)} \cdot \mW \vx_j^{(l)}),
\end{equation} 
where $\vx_j^{(l)}$ is the hidden feature for node $j$ in the $l$-th layer, $\mW$ is a weight matrix, and $\sigma$ is a nonlinear function such as $\mathrm{ReLU}$.
Given two nodes $i$ and $j$ in $\mathcal{G}$, their edge weights are formulated as,
\begin{equation}\label{eq:edge_weight}
	\small
	\alpha_{ij}^{(l)}=\frac{A_{ij}\exp(\mathrm{LReLU}(\va_{(l)}^T[\mW\vx_i^{(l)} \Vert \mW\vx_j^{(l)}]))}{\sum_{k \in \mathcal{N}_i}{A_{ik}\exp(\mathrm{LReLU}(\va_{(l)}^T[\mW\vx_i^{(l)} \Vert \mW\vx_k^{(l)}]))}}
\end{equation} 
where $\mathcal{N}_i$ represents the neighborhood of node $i$ in the graph, $\va_{(l)}$ is a learnable weight vector, and $\Vert$ denotes the concatenation operation. 
In practice, we stack two graph attention layers to form our ProtoGR.
Moreover, we combine the hidden features with the input features in the last layer to enhance the discriminability, 
\begin{equation}\label{eq:concat}
\mX^{(L)}=\mX^{(L)} + \mX^{(0)}
\end{equation} 
To further strengthen the discriminability of class prototypes, we perform node classification based on $\mathcal{G}$. 
To be specific, the last layer of ProtoGR predicts the class label, which can be formulated as, 
\begin{equation}\label{eq:pc}
	\hat{y}=\softmax({\text{FC}}(\text{ProtoGR}(x,\mathcal{G}))),
\end{equation} 
where $x$ is the node features, $\hat{y}$ denotes the predicted label, and FC denotes a fully-connected layer.
Finally, the node classification loss of ProtoGR module is defined as $\mathcal{L}_{\text{ProtoGR}}$.

\subsubsection{Prototypical Category-aware Contrastive Learning~(ProtoCCL)}
In addition to the topological relations among prototypes learned by ProtoGR, we further encode the semantic structures into the embedding space by introducing a contrastive learning objective for the task of category discrimination.

Contrastive learning based methods~\cite{hadsell2006dimensionality,he2020momentum,li2021prototypical,wang2021exploring,hu2021region} have demonstrated compelling results in (self-supervised) representation learning. 
Among them, InfoNCE~\cite{oord2018representation} is a representative contrastive loss function, 
\begin{equation}\small\label{eq:InfoNCE}
	\!\!\!\!\mathcal{L}^{\text{InfoNCE}}_I\!=\!-\log\frac{\exp(\bm{v}\!\cdot\!\bm{v}^+/\tau)}{\exp(\bm{v}\!\cdot\!\bm{v}^+/\tau)
		\!+\!\sum_{\bm{v}^-\in \mathcal{N}_I}\exp(\bm{v}\!\cdot\!\bm{v}^-/\tau)}\!\!
\end{equation}
where $\bm{v}^+$ is positive embeddings for $I$,
$\mathcal{N}_I$ contains embeddings of negatives,
and $\tau$ is a temperature hyper-parameter. 

In ProtoCCL, we utilize the contrastive loss in a fully supervised way considering the presence of class labels.
For a prototype $\bm{c}$ (query) with its category label $k$, the positive samples $\bm{c}^+$ (positive keys) are other prototypes from class $k$, while the negatives samples $\bm{c}^-$ (negative keys) are prototypes from other classes.
Formally, ProtoCCL for a prototype from category $k$ can be formulated as follows,
\begin{equation}\small\label{eq:ProtoCCL}
	\!\!\!\!\mathcal{L}^{\text{ProtoCCL}}_k\!=\!\frac{1}{|\mathcal{C}_k|}\sum_{\bm{c}^+\in\mathcal{C}_k}\!\!-\log\frac{\exp(\bm{c}\!\cdot\!\bm{c}^+/\tau)}{\exp(\bm{c}\!\cdot\!\bm{c}^+/\tau)
		\!+\!\sum_{\bm{c}^-\in \mathcal{N}_k}\exp(\bm{c}\!\cdot\!\bm{c}^-/\tau)}\!\!
\end{equation}
where $\mathcal{C}_k$ and $\mathcal{N}_k$ denote prototype feature collections of the positive and negative samples, for prototype $\bm{c}$.

The proposed ProtoCCL shares the principle of contrastive loss~\cite{hadsell2006dimensionality} to output small value when prototype $\bm{c}$ is similar to its positive key and dissimilar to all other negative keys. 
It is noteworthy that the constructed memory bank only stores the prototypes instead of all samples, thus significantly reducing the computational costs. 

\subsection{Training Losses}
The overall training process can be divided into two stages. 
In the first stage, we obtain the pseudo domain labels of all source samples by optimizing Eq.~(\ref{eq:domain_prediction}) and cross-entropy loss $\mathcal{L}_\mathrm{cls}$. Formally, $\mathcal{L}_\mathrm{cls}$ is defined as follows,
\begin{equation}\label{eq:cls}
	\mathcal{L}_{\mathrm{cls}}=-\frac{1}{N}\sum_{i=1}^N\sum\limits_{j=1}^K\,\mathbbm{1}[y_i=j]\log({F_c\circ{G(x_i)}})
\end{equation} 
where $G$ denotes the feature extractor and $F_c$ represents the class label predictor.
In the second stage, by using the domain labels predicted in the first stage, we have the following loss for our \cm\ model,
\begin{equation}\label{eq:overall}
\mathcal{L}_{\text{COMEN}}=\mathcal{L}_{\text{cls}}+ \lambda\mathcal{L}_{\text{ProtoPR}} + \gamma\mathcal{L}_{\text{ProtoCCL}}
\end{equation}
where $\lambda$ and $\gamma$ is the trade-off parameters that balance different training losses.

\section{Experiments}
\subsection{Datasets and Evaluation Protocol}
We evaluate out-of-domain accuracy of \cm\ on four DG benchmarks: \textbf{PACS}~\cite{li2017deeper}, \textbf{Digits-DG}~\cite{zhou2020learning}, \textbf{VLCS}~\cite{fang2013unbiased}, and \textbf{Office-Home}~\cite{venkateswara2017deep}. 
(1) PACS, which includes 9,991 images of seven categories, is the most commonly used DG benchmark due to its large distributional shift across four domains (\emph{Photo}, \emph{Art Painting}, \emph{Cartoon}, and \emph{Sketch}). 
(2) Digits-DG, which is a handwritten digit recognition benchmark, is composed of four domains (\emph{MNIST}~\cite{lecun1998gradient}, \emph{MNIST-M}~\cite{ganin2015unsupervised}, \emph{SVHN}~\cite{netzer2011reading}, and \emph{SYN}~\cite{ganin2015unsupervised}) where the domain shifts stem from the variations of background, style, and color. 
(3) VLCS collects images from \emph{PASCAL VOC 2007}~\cite{everingham2010pascal}, \emph{LabelMe}~\cite{russell2008labelme}, \emph{Caltech}~\cite{fei2004learning}, and \emph{Sun}~\cite{xiao2010sun} and contains five categories in all. 
(4) Office-Home, which is an object recognition dataset in office and home environments, consists of 15,500 images of 65 categories from four domains (\emph{Artistic}, \emph{Clipart}, \emph{Product}, and \emph{Real World}).

To facilitate a fair comparison, we follow the leave-one-domain-out evaluation protocol as in~\cite{li2017deeper,carlucci2019domain,li2021simple}, \emph{i.e.,} one domain is chosen as the held out domain and the remaining domains are seen as source domains. 
For PACS, Digits-DG, and Office-Home, we follow the train and val splits established in~\cite{li2017deeper,zhou2020learning,zhou2021domain}. 
For VLCS, we follow the previous works~\cite{carlucci2019domain,dg_mmld} to split 30\% of the source samples as validation datasets.
In the testing phase, we use all target samples to compute the classification accuracy of the model that exhibits the best performance in the validation dataset. 
\subsection{Implementation Details}
For PACS and Office-Home, we use ResNet-18~\cite{he2016deep} pre-trained on the ImageNet as the backbone architecture.
For Digits-DG, we follow~\cite{zhou2020learning} to construct the backbone with four 64-kernel 3$\times$3 \emph{conv} layers and a softmax layer.
For VLCS, we use AlexNet~\cite{krizhevsky2012imagenet} pre-trained on the ImageNet as the backbone architecture.
We train the networks using SGD with momentum of 0.9 and weight decay of 5e-4 for 50 epochs. 
For PACS, VLCS, and Office-Home, the initial learning rate is set to 0.001, which is decayed by 0.1 at 20 epochs.
For Digits-DG, we train the network from scratch and the initial learning rate is set to 0.05, which is decayed by 0.1 every 20 epochs.
The batch size is set to 128 in all experiments. 
For Eq.~(\ref{eq:ProtoCCL}), we set $\tau=0.5$.
For Eq.~(\ref{eq:overall}), we set $\lambda=0.1$ and $\gamma=0.1$. 
We evaluate the performance in all classes and report the average accuracy over 3 runs with different random seeds.  
All experiments are implemented on the PyTorch framework using a 1080Ti GPU.

\begin{center}
	\begin{table}[!t]
		\caption{Domain Generalization results on PACS benchmark.}\label{tab:PACS}
		\centering
		\small
		\setlength\tabcolsep{5pt}
		\begin{tabular}{ccccc|a}
			\toprule
            Method & Art & Cartoon & Photo & Sketch & Avg \\
			\hline
			\multicolumn{6}{c}{\emph{w/ Domain Supervision}} \\
			\hline
			CCSA~\cite{motiian2017unified} & 80.5 & 76.9 & 93.6 & 66.8 & 79.4 \\
			MMD-AAE~\cite{li2018domain} & 75.2 & 72.7 & 96.0 & 64.2 & 77.0 \\
			CrossGrad~\cite{shankar2018generalizing} & 79.8 & 76.8 & 96.0 & 70.2 & 80.7 \\
			Metareg~\cite{balaji2018metareg} & \underline{83.7} & 77.2 & 95.5 & 70.3 & 81.7 \\
			MASF~\cite{dou2019domain} & 80.3 & 77.2 & 95.0 & 71.7 & 81.1 \\
			Epi-FCR~\cite{li2019episodic} & 82.1 & 77.0 & 93.9 & 73.0 & 81.5 \\
			L2A-OT~\cite{zhou2020learning} & 83.3 & 78.2 & \textbf{96.2} & 73.6 & 82.8 \\
			SagNet~\cite{nam2021reducing} & 83.6 & 77.7 & 95.5 & 76.3 & 83.3 \\
			SelfReg~\cite{kim2021selfreg} & 82.3 & 78.4 & \textbf{96.2} & \underline{77.5} & 83.6 \\
			MixStyle~\cite{zhou2021domain} & \textbf{84.1} & \underline{78.8} & \underline{96.1} & 75.9 & \underline{83.7} \\
			\hline
			\multicolumn{6}{c}{\emph{w/o Domain Supervision}} \\
			\hline
			DeepAll~\cite{li2019episodic} & 77.6 & 73.9 & 94.4 & 70.3 & 79.1 \\
			JiGen~\cite{carlucci2019domain} & 79.4 & 75.3 & 96.0 & 71.6 & 80.5 \\
			MMLD~\cite{dg_mmld} & 81.3 & 77.2 & \underline{96.1} & 72.3 & 81.7 \\
			RSC~\cite{huang2020self} & 79.9 & 76.9 & 94.6 & 77.1 & 81.1 \\
			EISNet~\cite{wang2020learning} & 81.9 & 76.4 & 96.0 & 74.3 & 82.2 \\
			SFA-A~\cite{li2021simple} & 81.2 & 77.8 & 93.9 & 73.7 & 81.7 \\
			SagNet~\cite{nam2021reducing} & 83.6 & 77.7 & 95.5 & 76.3 & 83.3 \\
			\hline
			\cm\ (Ours) & 82.6 & \textbf{81.0} & 94.6 & \textbf{84.5} & \textbf{85.7} \\
			\bottomrule
		\end{tabular}
	\end{table}
\end{center}

\begin{center}
	\begin{table}[t]
		\caption{Domain Generalization results on Digits-DG benchmark.}\label{tab:Digits-DG}
		\vspace{-0.2cm}
		\centering
		\small
		\setlength\tabcolsep{3pt}
		\begin{tabular}{ccccc|a}
			\toprule
			Method & MNIST & MNIST-M & SVHN & SYN & Avg \\
			\hline
			\multicolumn{6}{c}{\emph{w/ Domain Supervision}} \\
			\hline
			CCSA~\cite{motiian2017unified} & 95.2 & 58.2 & 65.5 & 79.1 & 74.5 \\
			MMD-AAE~\cite{li2018domain} & 96.5 & 58.4 & 65.0 & 78.4 & 74.6 \\
			CrossGrad~\cite{shankar2018generalizing} & \underline{96.7} & 61.1 & 65.3 & 80.2 & 75.8 \\
			DDAIG~\cite{zhou2020deep} & 96.6 & 64.1 & 68.6 & 81.0 & 77.6 \\
			L2A-OT~\cite{zhou2020learning} & 96.7 & 63.9 & 68.6 & 83.2 & 78.1 \\
			MixStyle~\cite{zhou2021domain} & 96.5 & 63.5 & 64.7 & 81.2 & 76.5 \\
			\hline
			\multicolumn{6}{c}{\emph{w/o Domain Supervision}} \\
			\hline
			DeepAll~\cite{zhou2020learning} & 95.8 & 58.8 & 61.7 & 78.6 & 73.7 \\
			JiGen~\cite{carlucci2019domain} & 96.5 & 61.4 & 63.7 & 74.0 & 73.9 \\
			SFA-A~\cite{li2021simple} & 96.5 & \underline{66.5} & \underline{70.3} & \underline{85.0} & \underline{79.6} \\
			\hline
			\cm\ (Ours) & \textbf{97.1} & \textbf{67.6} & \textbf{75.1} & \textbf{91.3} & \textbf{82.3} \\
			\bottomrule
		\end{tabular}
	\end{table}
\end{center}
\vspace{-0.5cm}
\begin{center}
	\begin{table}[t]
		\caption{Domain Generalization results on VLCS benchmark.}\label{tab:VLCS}
		\vspace{-0.2cm}
		\centering
		\small
		\setlength\tabcolsep{7pt}
		\begin{tabular}{ccccc|a}
			\toprule
            Method & V & L & C & S & Avg \\
            \hline
			\multicolumn{6}{c}{\emph{w/ Domain Supervision}} \\
            \hline
			D-MTAE~\cite{ghifary2015domain} & 63.9 & 60.1 & 89.1 & 61.3 & 68.6 \\
			CCSA~\cite{motiian2017unified} & 67.1 & 62.1 & 92.3 & 59.1 & 70.2 \\
			DBA-DG~\cite{li2017deeper} & 70.0 & 63.5 & 93.6 & 61.3 & 72.1 \\
			CIDDG~\cite{li2018deep} & 64.4 & 63.1 & 88.8 & 62.1 & 69.6 \\
			MMD-AAE~\cite{li2018domain} & 67.7 & 62.6 & 94.4 & 64.4 & 72.3 \\
			MLDG~\cite{li2018learning} & 67.7 & 61.3 & 94.4 & 65.9 & 72.3 \\
			Epi-FCR~\cite{li2019episodic} & 67.1 & \underline{64.3} & 94.1 & 65.9 & 72.9 \\
			MASF~\cite{dou2019domain} & 69.1 & \textbf{64.9} & 94.8 & 67.6 & \underline{74.1} \\
			\hline
			\multicolumn{6}{c}{\emph{w/o Domain Supervision}} \\
			\hline
			DeepAll~\cite{li2021simple} & 71.9 & 59.2 & 96.9 & 62.6 & 72.7 \\
			JiGen~\cite{carlucci2019domain} & 70.6 & 60.9 & 96.9 & 64.3 & 73.2 \\
			MMLD~\cite{dg_mmld} & \underline{72.0} & 58.8 & 96.7 & \textbf{68.1} & 73.9 \\
			SFA-A~\cite{li2021simple} & 70.4 & 62.0 & \textbf{97.2} & 66.2 & 74.0 \\
			\hline
			\cm\ (Ours) & \textbf{72.8} & 62.6 & \underline{97.0} & \underline{67.6} & \textbf{75.0} \\
			\bottomrule
		\end{tabular}
	\end{table}
\end{center}
\vspace{-0.5cm}
\begin{center}
	\begin{table}[t]
		\caption{Domain Generalization results on Office-Home.}\label{tab:Office-Home}
		\vspace{-0.2cm}
		\centering
		\small
		\setlength\tabcolsep{5pt}
		\begin{tabular}{ccccc|a}
			\toprule
			Method & Art & Clipart & Product & Real & Avg \\
			\hline
			\multicolumn{6}{c}{\emph{w/ Domain Supervision}} \\
			\hline
			CCSA~\cite{motiian2017unified} & 59.9 & 49.9 & 74.1 & 75.7 & 64.9 \\
			MMD-AAE~\cite{li2018domain} & 56.5 & 47.3 & 72.1 & 74.8 & 62.7 \\
			CrossGrad~\cite{shankar2018generalizing} & 58.4 & 49.4 & 73.9 & 75.8 & 64.4 \\
			DDAIG~\cite{zhou2020deep} & 59.2 & 52.3 & 74.6 & 76.0 & 65.5 \\
			DOSN~\cite{seo2020learning} & 59.4 & 45.7 & 71.8 & 74.7 & 62.9 \\
			L2A-OT~\cite{zhou2020learning} & \textbf{60.6} & 50.1 & \underline{74.8} & \textbf{77.0} & \underline{65.6} \\
			MixStyle~\cite{zhou2021domain} & 58.7 & \underline{53.4} & 74.2 & 75.9 & 65.5 \\
			\hline
			\multicolumn{6}{c}{\emph{w/o Domain Supervision}} \\
			\hline
			DeepAll~\cite{zhou2020learning} & 58.9 & 49.4 & 74.3 & 76.2 & 64.7 \\
			JiGen~\cite{carlucci2019domain} & 53.0 & 47.5 & 71.5 & 72.8 & 61.2 \\
			RSC~\cite{huang2020self} & 58.4 & 47.9 & 71.6 & 74.5 & 63.1 \\
			SagNet~\cite{nam2021reducing} & \underline{60.2} & 45.4 & 70.4 & 73.4 & 62.4 \\
			\hline
			\cm\ (Ours) & 57.6 & \textbf{55.8} & \textbf{75.5} & \underline{76.9} & \textbf{66.5} \\
			\bottomrule
			\vspace{-0.5cm}
		\end{tabular}
	\end{table}
\end{center}
\vspace{-3cm}
\subsection{Comparisons with State-of-the-Arts}
\paragraph{Comparison Methods.} We extensively compare the proposed \cm\ against state-of-the-art DG methods, which can be classified into two groups: \textbf{w/ Domain Supervision} and \textbf{w/o Domain Supervision}. 
\emph{w/ Domain Supervision} indicates that these methods relies on the domain information to implement their algorithms.
Note that \emph{w/o Domain Supervision} does not imply that the included methods also explicitly discover and model the latent domains.
By contrast, they usually treat the compound source domains as a whole.
For all the compared approaches, we summarize the domain generalization results reported in their original papers.
\textbf{DeepAll} represents the standard supervised learning by directly combining all source domains.

\textbf{PACS and Digits-DG.} The domain generalization results on PACS and Digits-DG benchmarks are reported in Table~\ref{tab:PACS} and Table~\ref{tab:Digits-DG}, 
we can observe that the proposed \cm\ outperforms all the baseline methods by a large margin and improves over state-of-the-art results by +2.0\%~(PACS) and 2.7\%~(Digits-DG) on average, clearly demonstrating that \cm\ is capable of extracting generalizable and transferable representations for effective out-of-distribution generalization.
In particular, \cm\ largely improves the classification accuracy on the hard generalization tasks, such as \emph{Sketch} (70.3\%~$\rightarrow$~84.5\%), \emph{MNIST-M} (58.8\%~$\rightarrow$~67.6\%), and \emph{SVHN} (61.7\%~$\rightarrow$~75.1\%), where the source domains are significantly different from the unseen target domain. This encouraging results further reveal the significance of discovering and modeling latent domains. 

\textbf{VLCS and Office-Home.} 
Table~\ref{tab:VLCS} and Table~\ref{tab:Office-Home} display the results on VLCS and Office-Home.
Our method consistently and substantially outperforms all the compared methods on most tasks, which indicates that \cm\ is versatile and scalable across different domain generalization scenarios. However, we notice that the improvements on VLCS are smaller than the results on the other datasets. The rationale is that the domain shifts of the VLCS dataset stems from the type of camera, and thus the visual appearances are more analogous to each other.
In addition, all methods including ours achieve less improvements on Office-Home dataset compared to the vanilla DeepAll baseline. 
The DeepAll model even outperforms the state-of-the-art \emph{w/o Domain Supervision} methods, such as RSC and SagNet.
By comparison, \cm\ significantly exceeds the performance of these methods on this scenario.

These experimental results also reveal two important observations. 
(1) Mixstyle~\cite{zhou2021domain} and SFA-A~\cite{li2021simple}, which introduce explicit feature augmentation strategies, is clearly better than the previous works that focus on feature alignment, such as MASF~\cite{dou2019domain} and MMLD~\cite{dg_mmld}.
(2) \cm\ substantially outperforms the state-of-the-art methods (Mixstyle and SFA-A), which demonstrates the superiority of \cm\ by estimating the domain assignment of each source sample and encoding semantic structure into latent space in virtue of prototypical representations.

\begin{center}
	\begin{table*}[htb]
		\caption{Ablation of \cm\ on four DG benchmarks (\%).}
		\vspace{-0.2cm}
		\centering
		\label{tab:ablation}
		\begin{tabular}{ccc|cccc|c}
			\toprule
			DSNorm & ProtoGR & ProtoCCL & PACS & Digits-DG & VLCS & Office-Home & Average\\
			\hline
			-                             &-         &-         & 79.1 & 73.7 & 72.7 & 64.7 & 72.6 \\
			\hline
			
			$ \checkmark$ &-         &-         & 81.3 & 75.4 & 73.4 & 65.0 & 73.8 \\
			-                             & $\checkmark$   &-         & 82.0 & 75.6 & 74.1 & 65.4 & 74.3 \\
			-                             &-         & $\checkmark$   & 81.3 & 76.1 & 73.5 & 65.1 & 74.0 \\
			-                             & $\checkmark$   & $\checkmark$   & 82.6 & 77.8 & 74.4 & 65.6 & 75.1 \\
			$\checkmark$                       & $\checkmark$   &-         & 83.5 & 79.0 & 74.3 & 65.8 & 75.7 \\
			$\checkmark$                      &-         & $\checkmark$   & 84.0 & 79.3 & 74.2 & 66.3 & 76.0 \\
			\hline
			$\checkmark$                       & $\checkmark$   & $\checkmark$   & 85.7 & 82.3 & 75.0 & 66.5 & 77.4 \\
			\bottomrule
			\vspace{-0.8cm}
		\end{tabular}
	\end{table*}
\end{center}

\begin{figure*}[htb]
	\centering
	\subfloat[DeepAll]{
		\centering
		\includegraphics[width=0.2\linewidth]{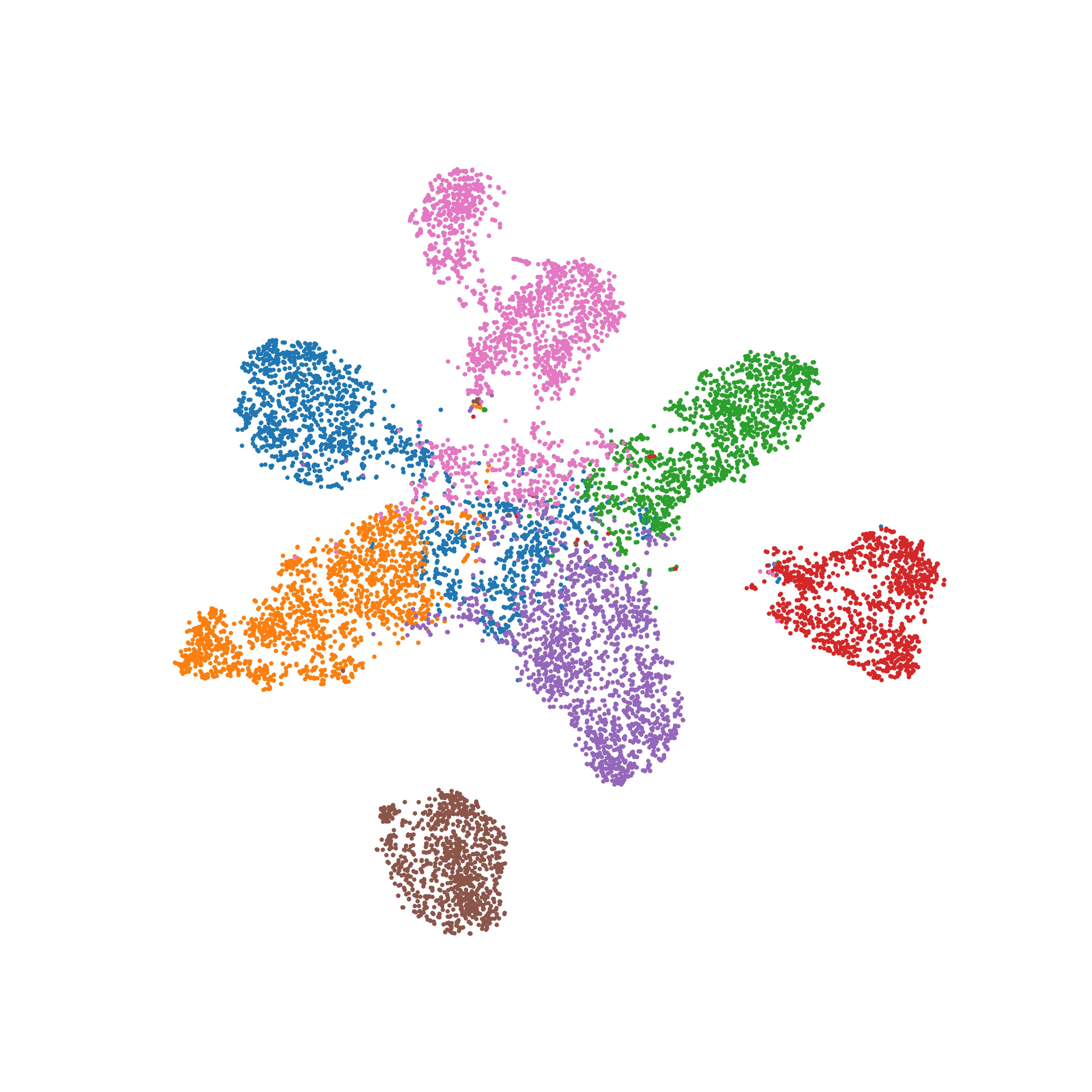}}
	\subfloat[\cm]{
		\centering
		\includegraphics[width=0.2\linewidth]{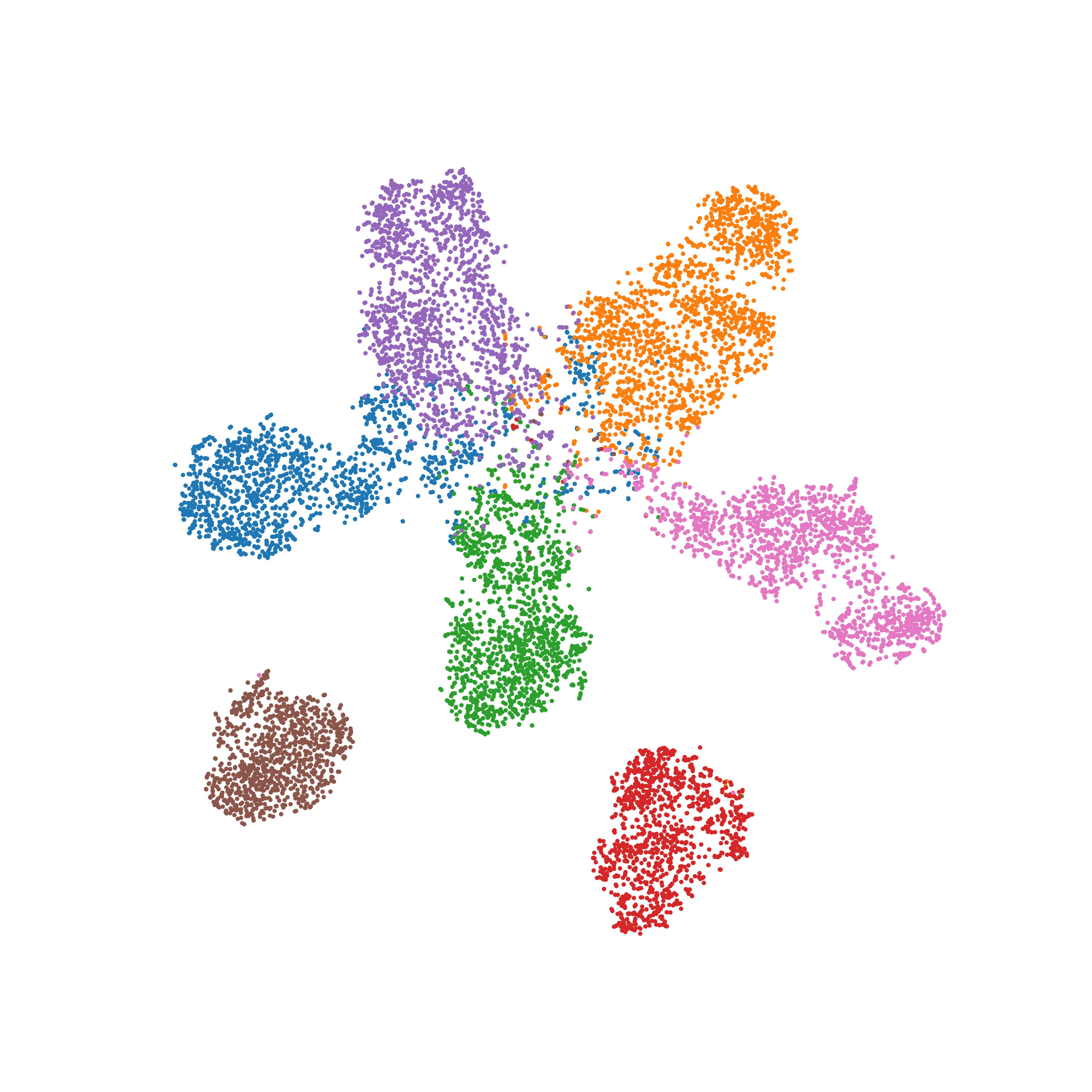}}
	\subfloat[DeepAll]{
		\centering
		\includegraphics[width=0.2\linewidth]{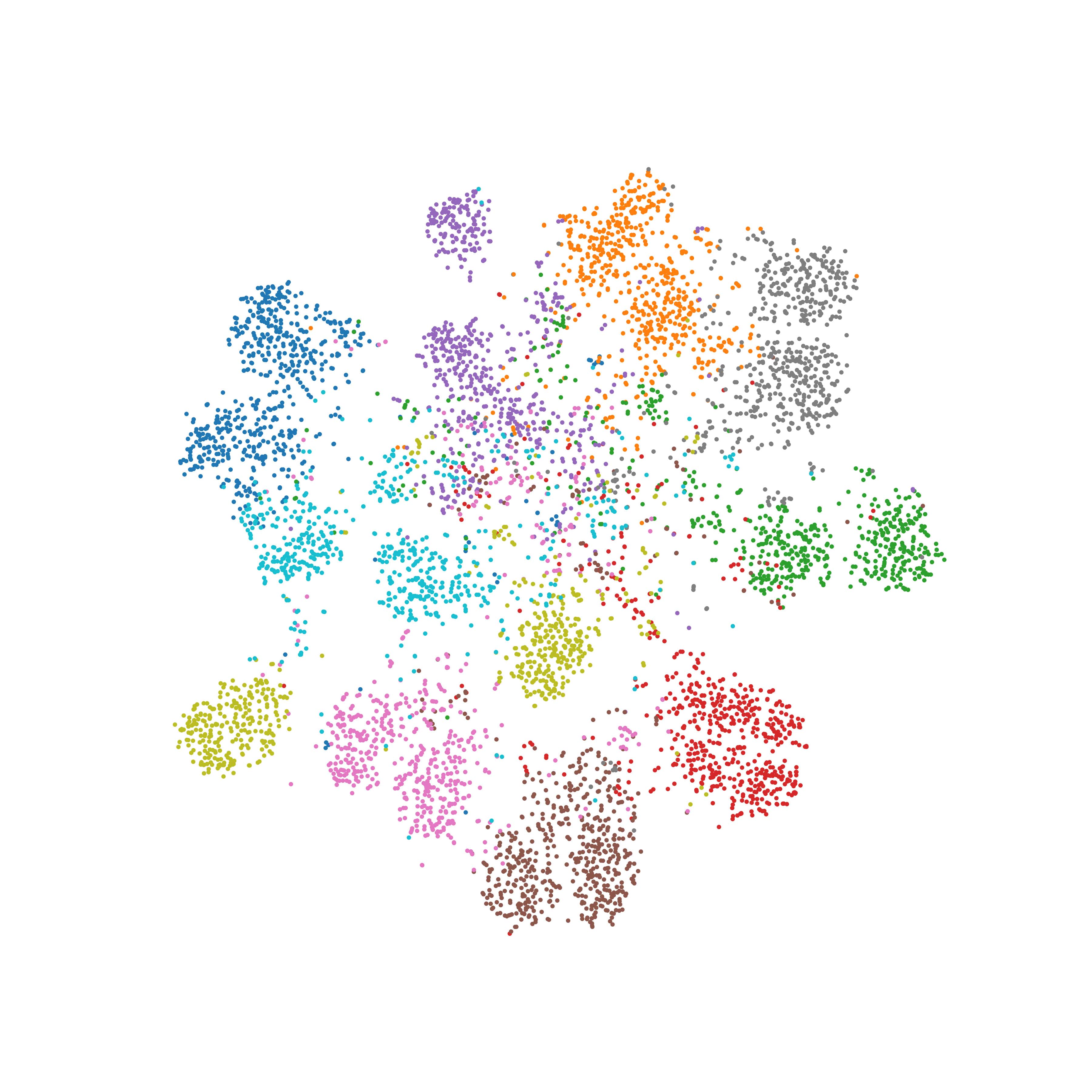}}
	\subfloat[\cm]{
		\centering
		\includegraphics[width=0.2\linewidth]{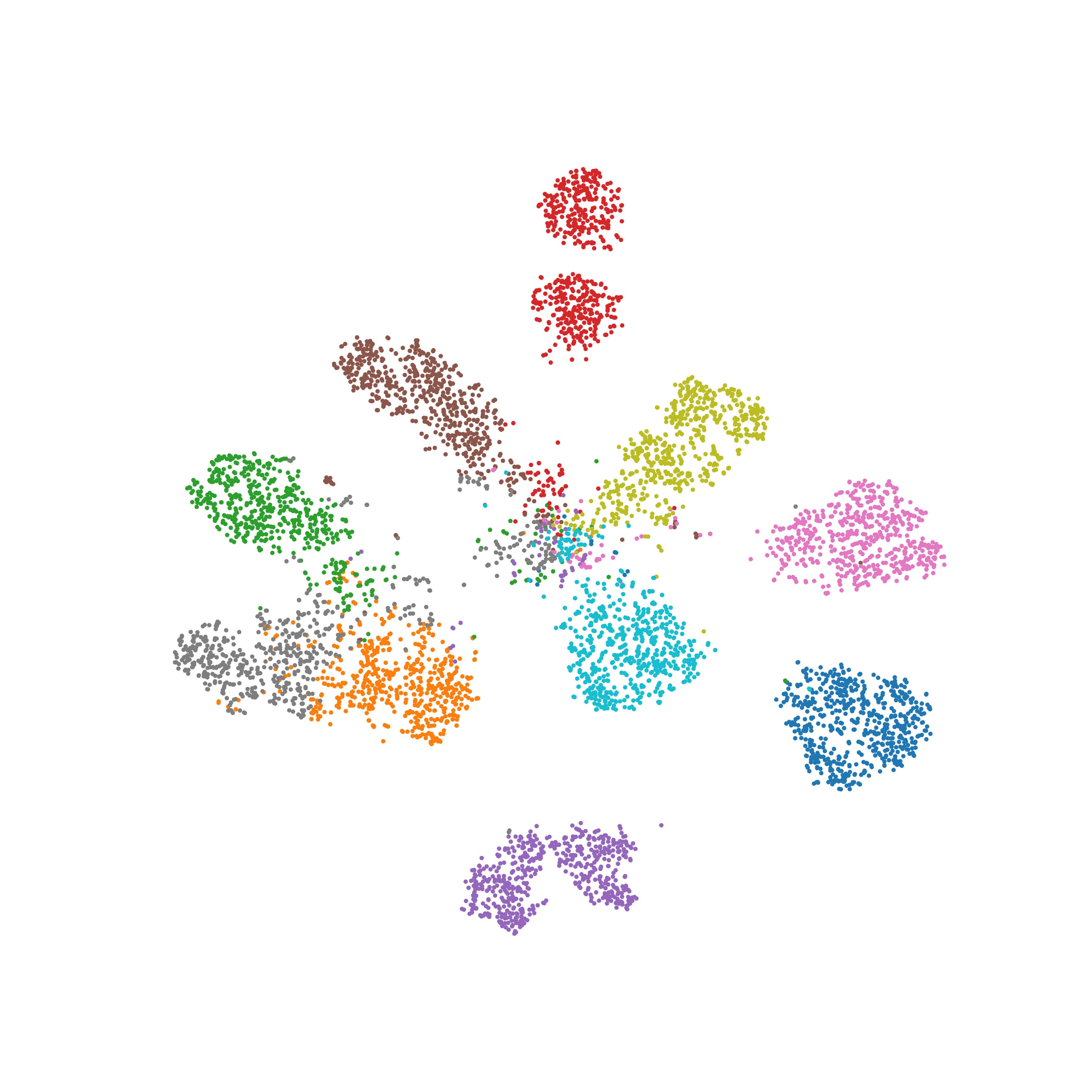}}
	\caption{The t-SNE visualization of deep network activations extracted by DeepAll and \cm\ on the PACS and Digits-DG datasets.}
	\vspace{-0.5cm}
	\label{fig:t-sne}
\end{figure*}

\begin{figure*}[!h]
	\centering
	\subfloat[DeepAll]{
	\centering
	\includegraphics[width=0.12\linewidth]{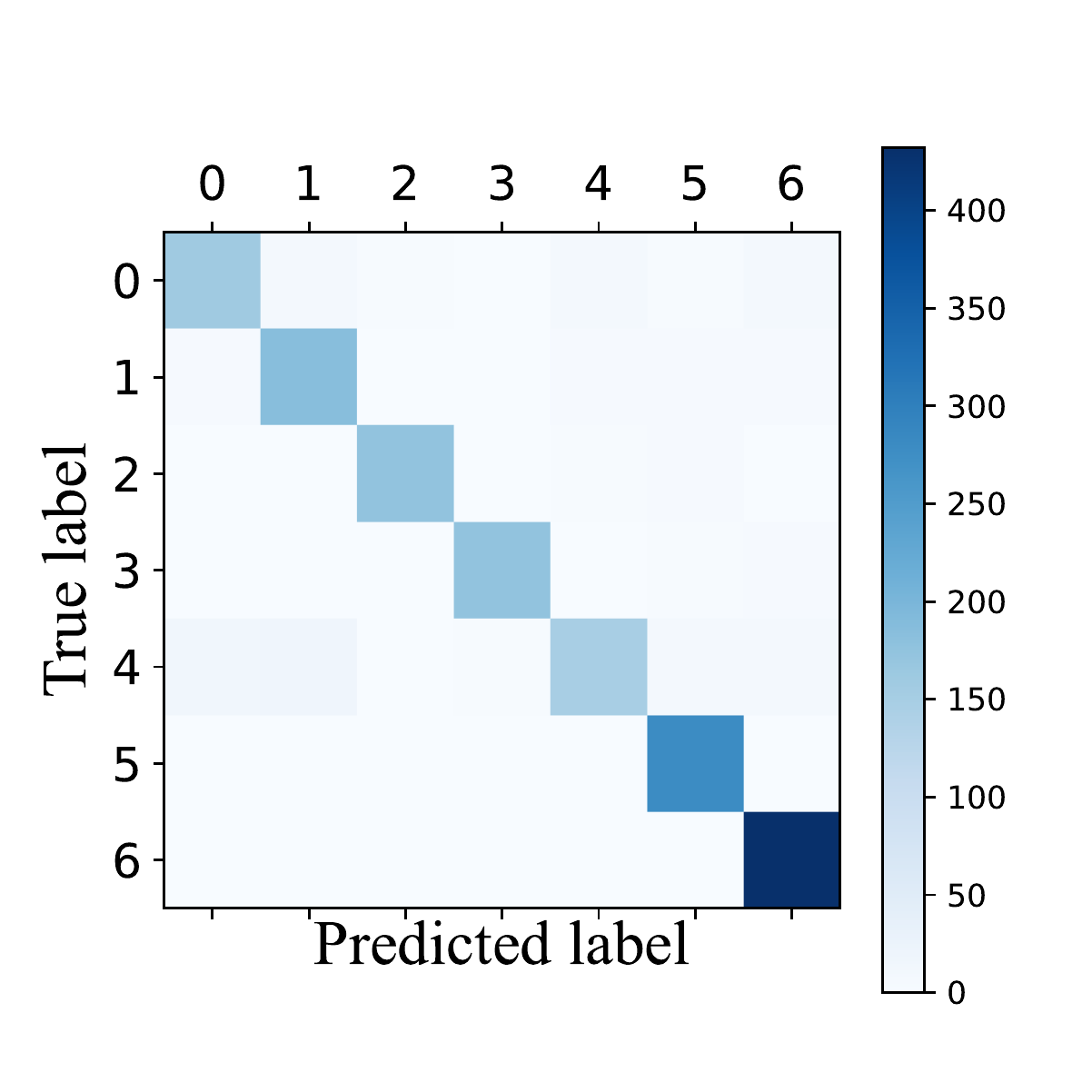}}
    \subfloat[\cm]{
	\centering
	\includegraphics[width=0.12\linewidth]{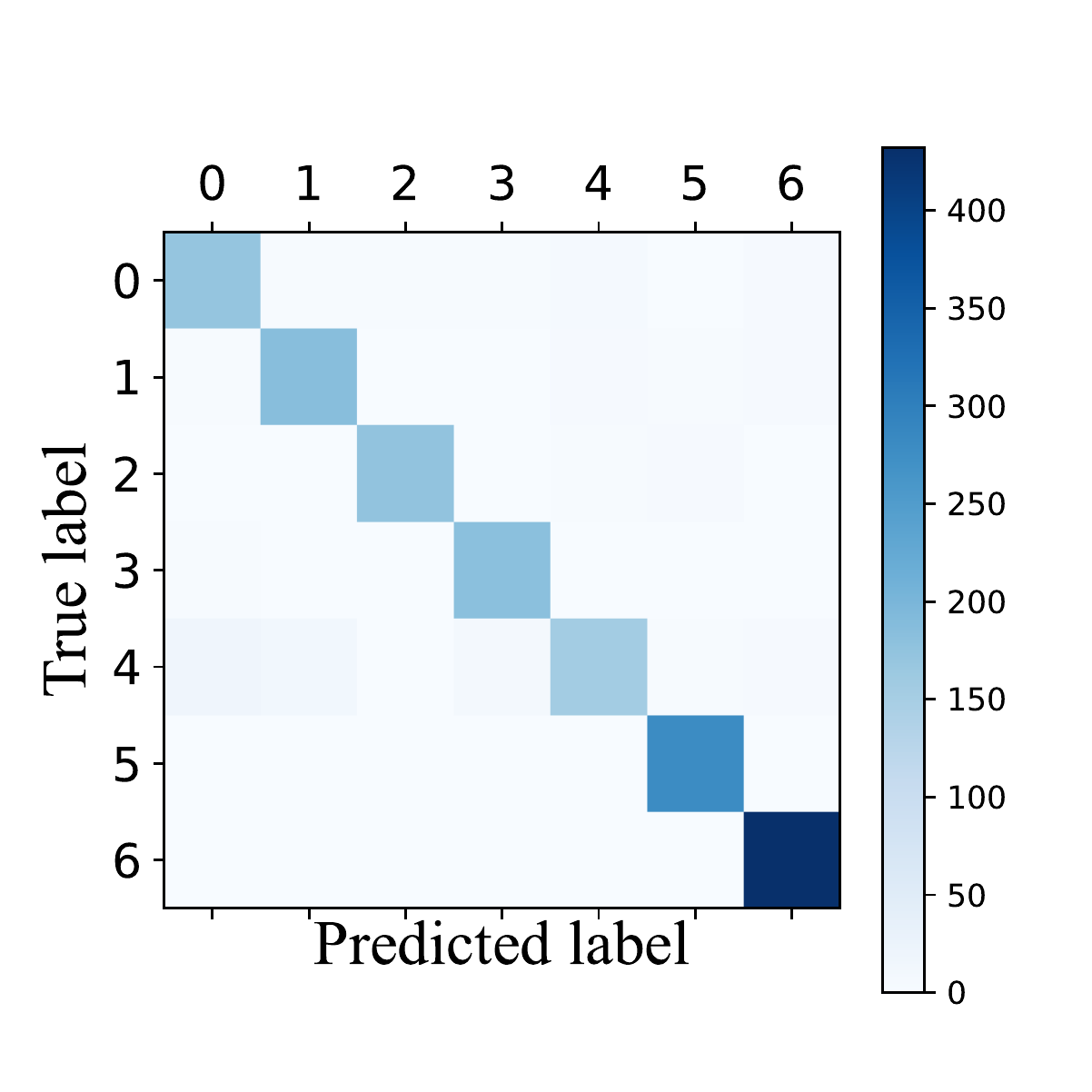}}
    \subfloat[DeepAll]{
	\centering
	\includegraphics[width=0.12\linewidth]{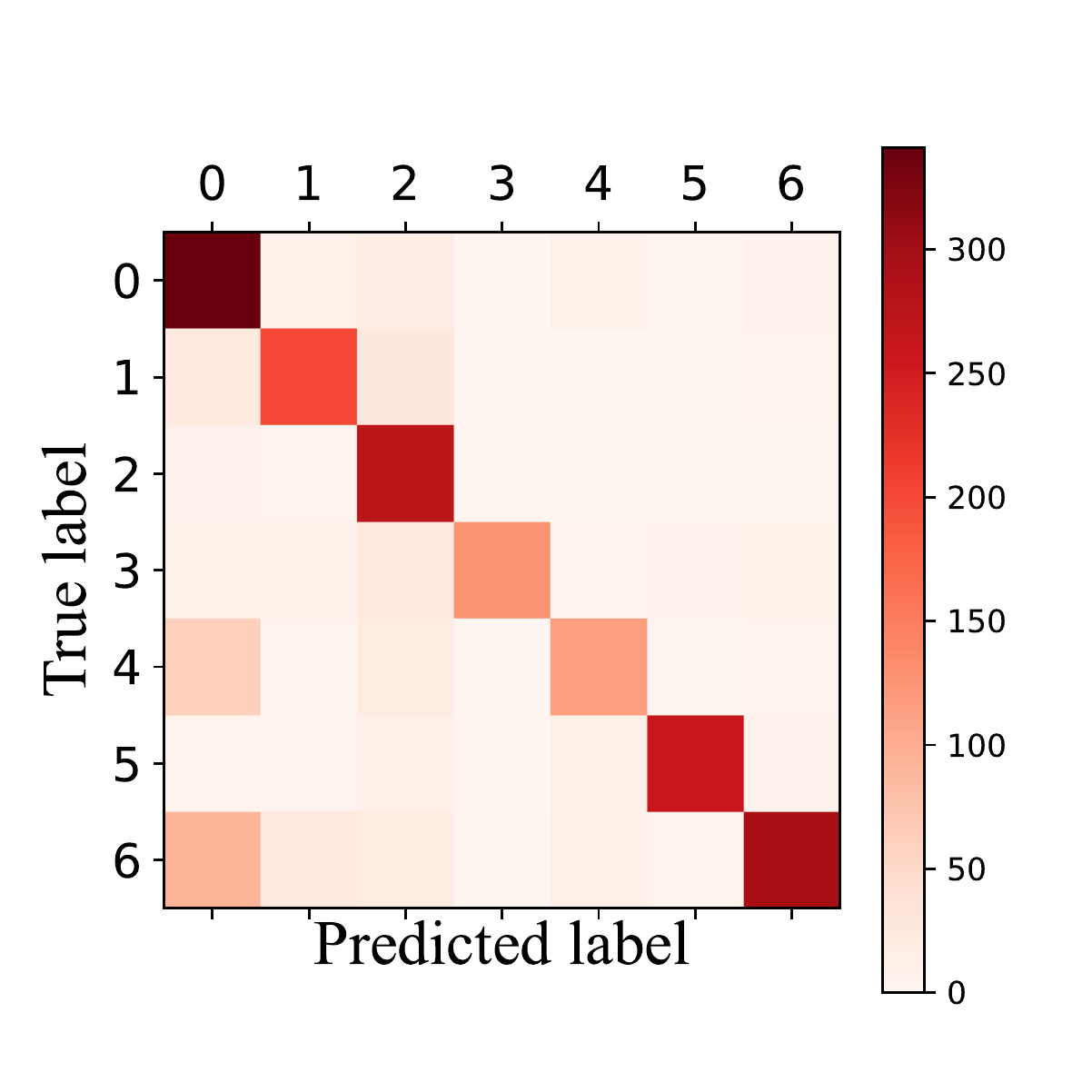}}
	\subfloat[\cm]{
	\centering
	\includegraphics[width=0.12\linewidth]{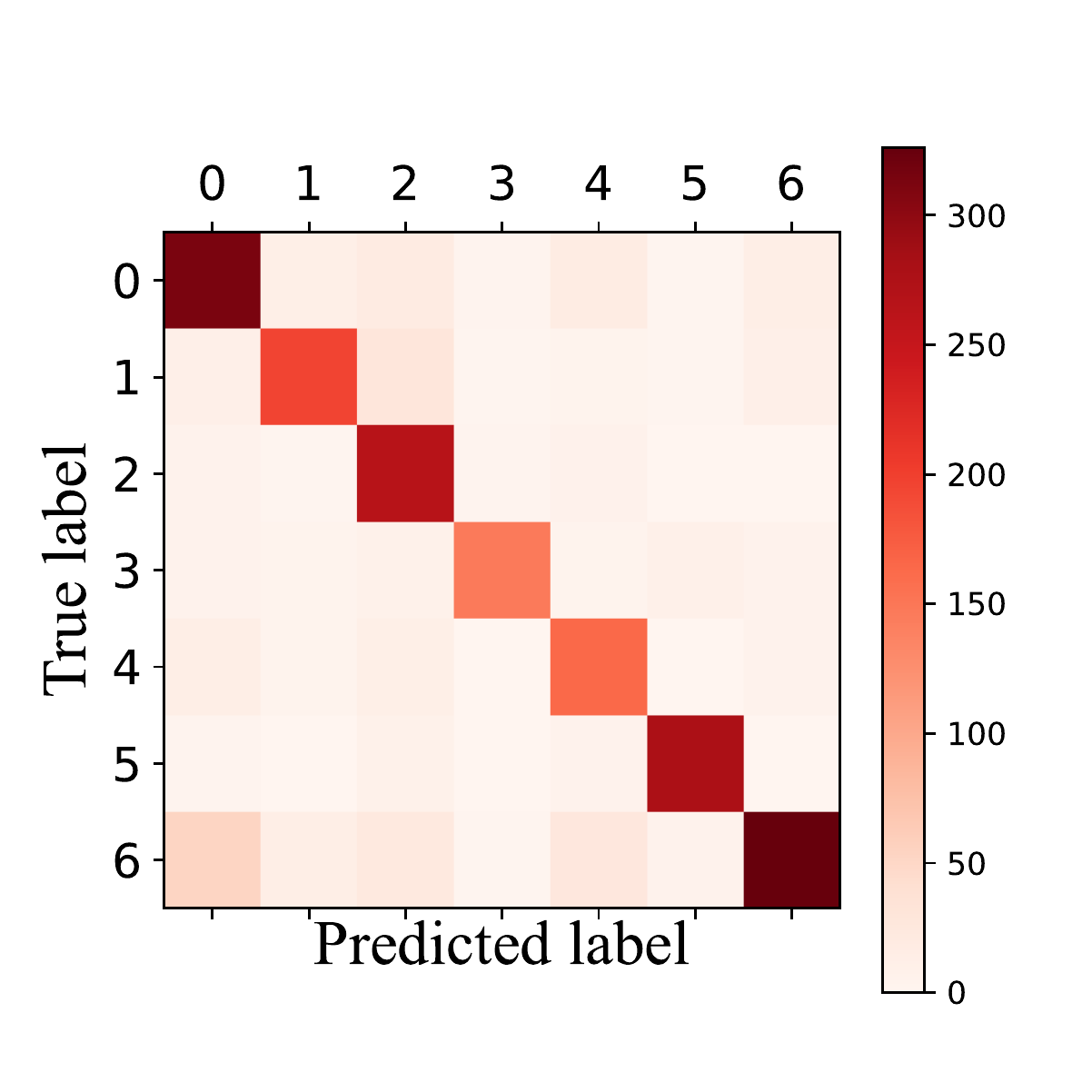}}
	\subfloat[DeepAll]{
		\centering
		\includegraphics[width=0.12\linewidth]{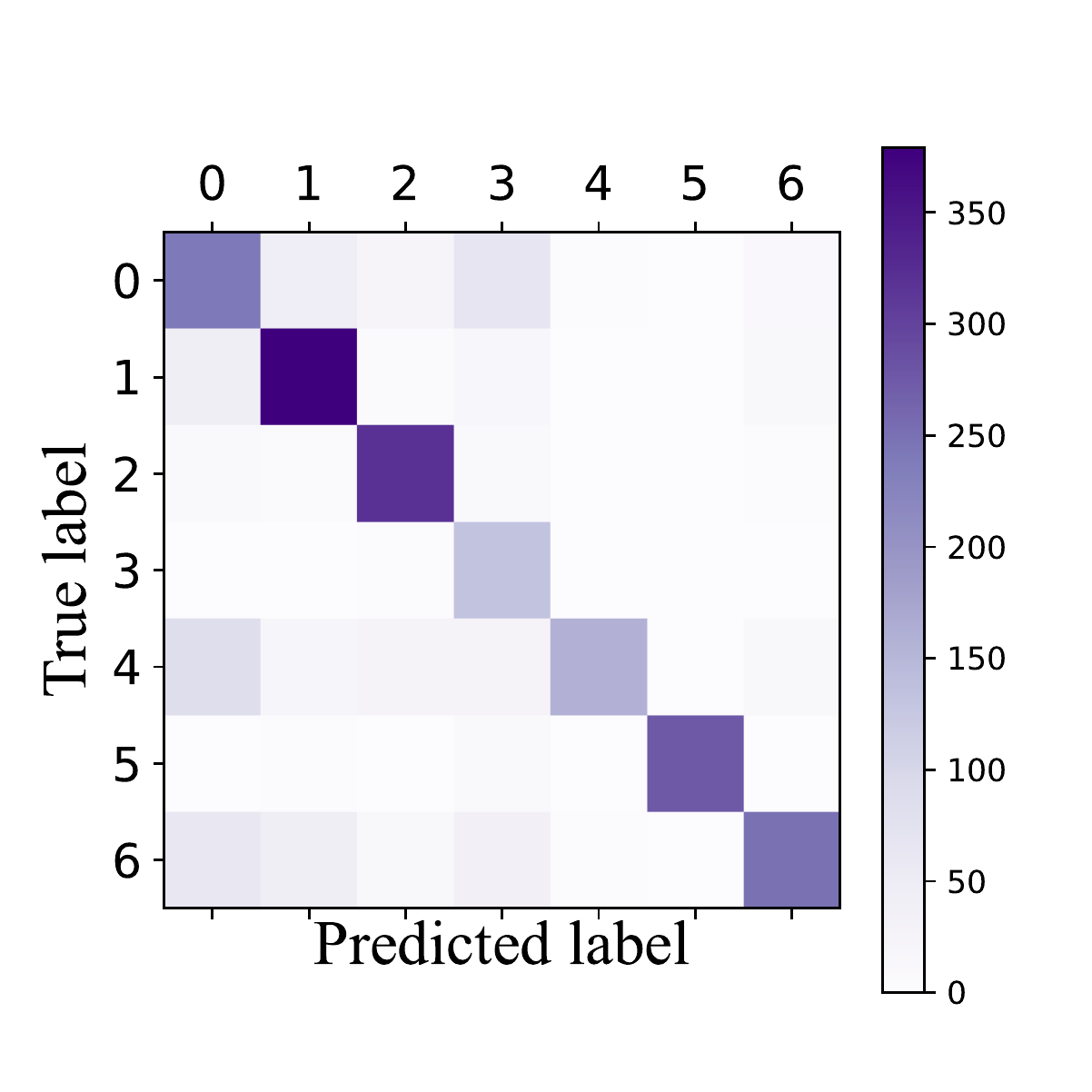}}
	\subfloat[\cm]{
		\centering
		\includegraphics[width=0.12\linewidth]{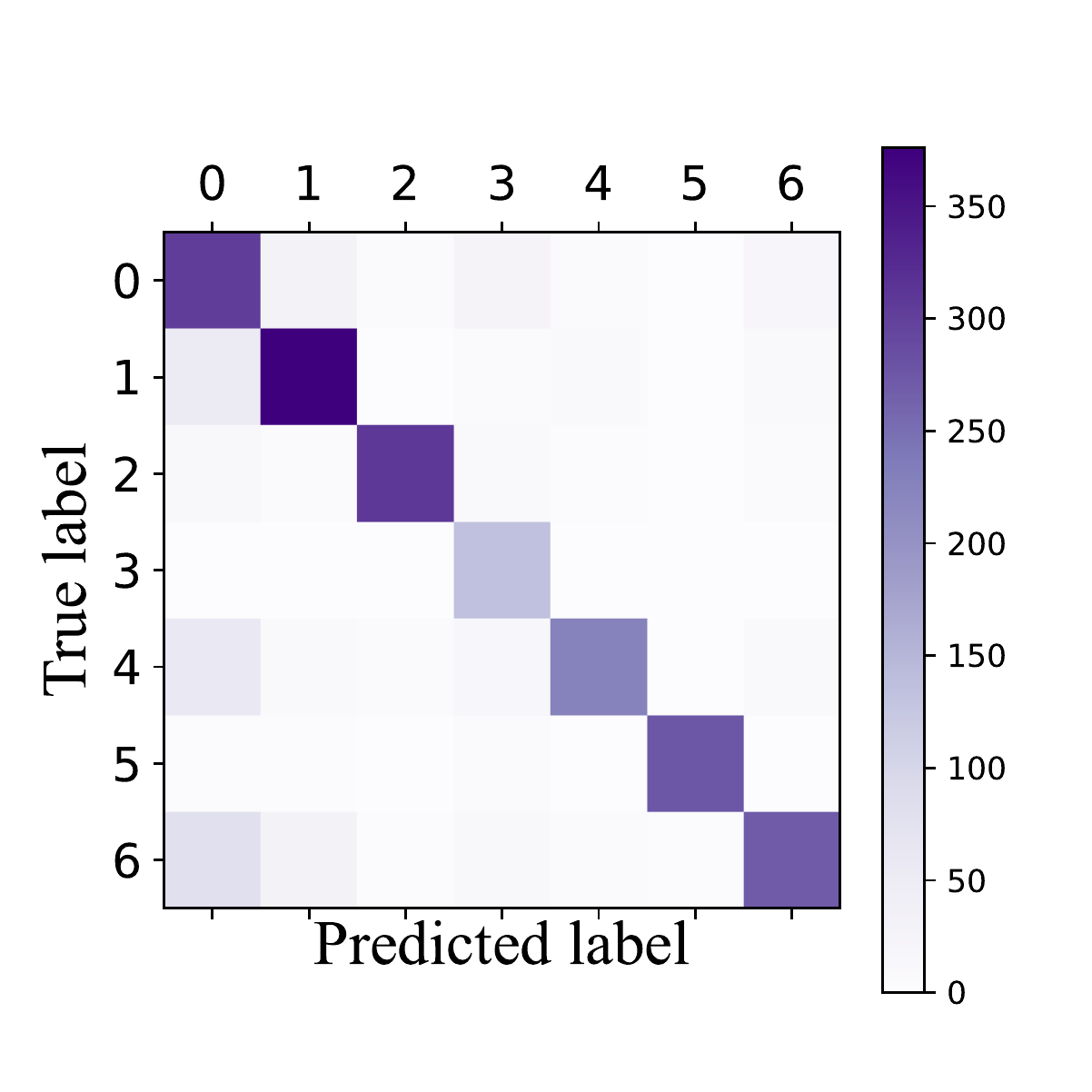}}
	\subfloat[DeepAll]{
		\centering
		\includegraphics[width=0.12\linewidth]{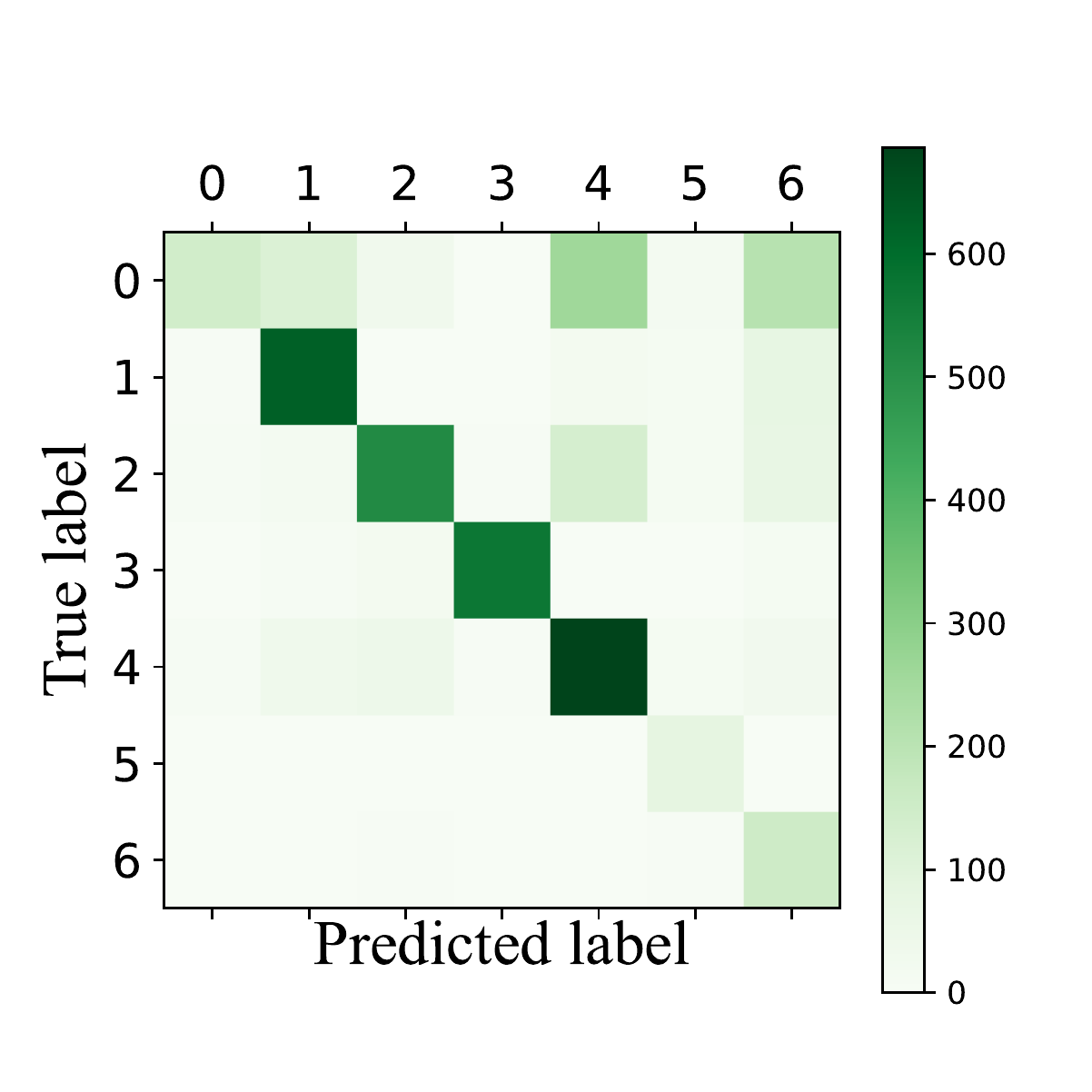}}
	\subfloat[\cm]{
		\centering
		\includegraphics[width=0.12\linewidth]{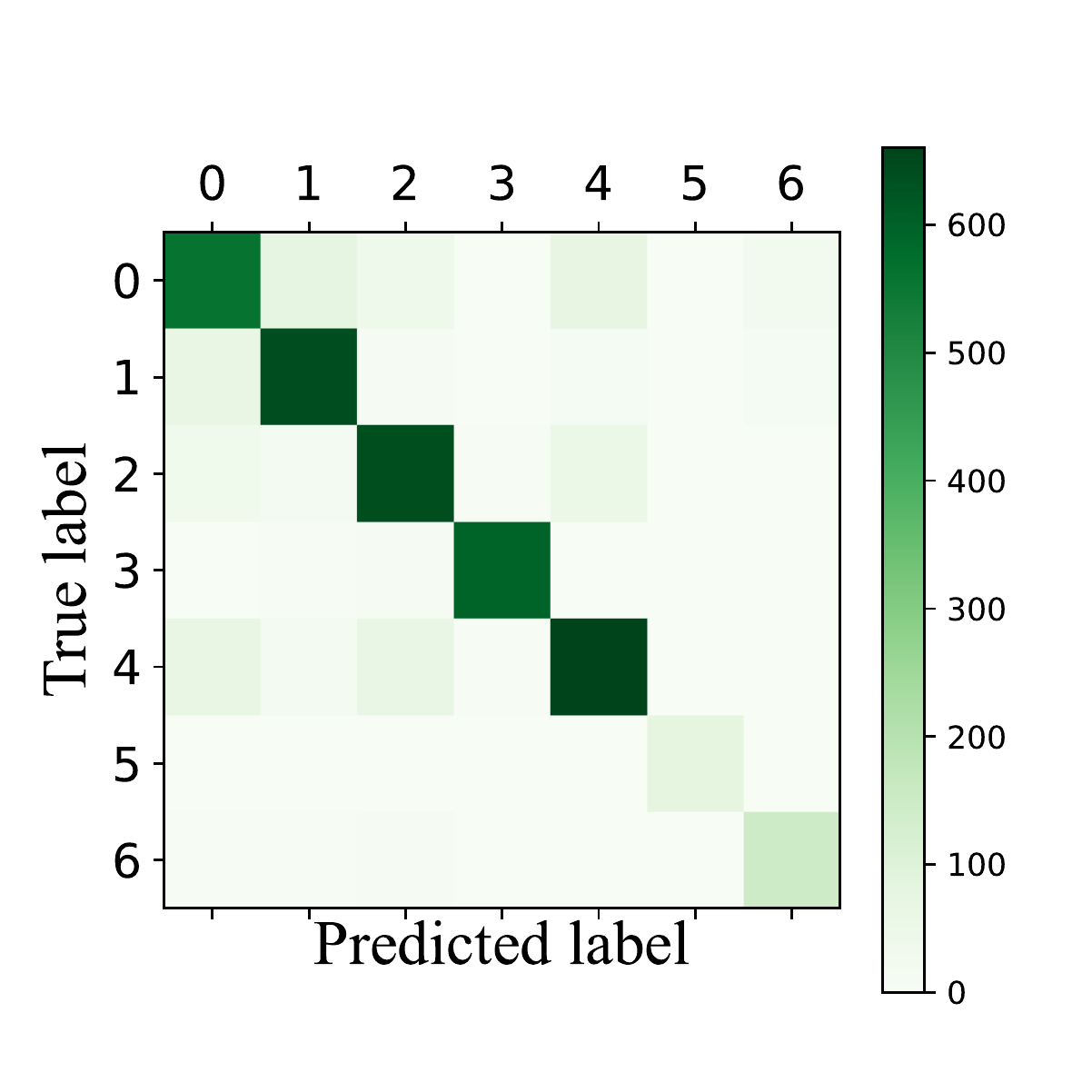}}
	\caption{The confusion matrices of DeepAll and \cm\ on PACS dataset. From left to right: Photo, Art, Cartoon, and Sketch.}
	\label{fig:confusion_matrix}
\end{figure*}
\vspace{-0.8cm}
\subsection{Further Empirical Analysis}
\textbf{Ablation Study.}
To study the relative contribution of each component in the proposed \cm, we perform an extensive ablation analysis on four DG benchmarks. The results are reported in Table~\ref{tab:ablation}, where the first raw corresponds to the DeepAll baseline. 
Specifically, we evaluate all possible combinations of the three proposed components: DSNorm, ProtoGR, and ProtoCCL. We can observe that all the components are designed reasonably and when any one of these components is removed, the performance drops accordingly. 
Noting that removing the DSNorm module leads to significant performance degeneration, which verifies the importance of discovering the latent domains in compound DG. Combining both ProtoGR and ProtoCCL, which model the prototypical relations in the feature space, shows clear improvement over DeepAll and DSNorm only.
For methods without DSNorm, we randomly and uniformly divide the compound source domains into multiple discrete domains so as to conduct the ProtoGR or/and ProtoCCL modules. In this case, we can observe that these two modules are robust to the noisy domain assignment results.

\textbf{Feature Visualization.}
We utilize t-SNE~\cite{van2008visualizing} for the visualizations of the deep network activations learned by DeepAll and \cm\ on the PACS and Digits-DG datasets. The results are shown in Figure~\ref{fig:t-sne}, where different colors stand for different classes. The features learned by DeepAll on the mixture of source domains cannot be reasonably separated, and the decision boundaries of some classes are ambiguous in the feature space. 
By contrast, \cm\ learns discriminative representations for both source and unseen target domains by simultaneously enlarging inter-class distance and suppressing the intra-class dispersion. 

\textbf{Confusion matrices.}
Figure~\ref{fig:confusion_matrix} respectively plots confusion matrices of DeepAll and \cm\ regarding the PACS dataset. We can see that the false prediction results are largely reduced by \cm\ especially on the hard generalization task (Cartoon and Sketch), revealing the key importance of discovering and modeling latent domains. 

\vspace{-0.2cm}
\section{Conclusion}
We study a challenging yet practical domain generalization problem, namely compound domain generalization.
To solve this problem, we propose a novel \cm\ framework to automatically discover and model the distinct underlying domains with three newly proposed modules.
In particular, DSNorm re-normalizes the multi-modal distributions to obtain domain information.
ProtoGR and ProtoCCL model the prototypical relations by simultaneously encoding semantic structure of feature space and learning discriminative domain-invariant representations. 
Experiments reveal that \cm\ significantly outperforms state-of-the-art DG methods on four benchmark datasets.

\cm\ is tailored for the classification task, and cannot be simply extended to other vision tasks, such as object detection and semantic segmentation.

\noindent
\textbf{Acknowledgement.} This work was partially supported by National Key Research and Development Program of China (No.2020YFC2003902) and Hong Kong Research Grants Council through Research Impact Fund (Grant R-5001-18).

{\small
\bibliographystyle{ieee_fullname}
\bibliography{egbib}
}

\end{document}